\documentclass{article}
\usepackage{spconf,amsmath,graphicx}
\usepackage{booktabs}
\usepackage{multirow}
\usepackage{array}   % needed for the >{...}p{} column trick
\usepackage{xcolor}
\usepackage{amssymb}
\usepackage{url}

\newsavebox{\resultsbox}
\usepackage{enumitem}
\setlist{nosep, leftmargin=14pt}

\usepackage[numbers]{natbib}
\title{Ultra-Lightweight 3D Medical Image Segmentation via Knowledge Distillation in Label-Scarce Domains}
\name{Rahim Ullah$^{1}$, Hasan Abbas$^{2}$, Syed Waqar Nabi$^{1}$}
\address{$^{1}$School of Computing, University of Glasgow, Glasgow, United Kingdom \\
         $^{2}$James Watt School of Engineering, University of Glasgow, Glasgow, United Kingdom  \\
         rahim.ullah@glasgow.ac.uk, hasan.abbas@glasgow.ac.uk, syed.nabi@glasgow.ac.uk
        }
\begin{document}
%\ninept
%
\maketitle
\begin{abstract}
Mamba is a neural network architecture that reduces computational complexity, and has been shown to be effective for 3D medical image segmentation. However, existing models still remain computationally expensive. We propose compressing Mamba by distillation while achieving competitive performance. Specifically, we demonstrate compression of a 139M-parameter SegMamba-V2 teacher into a lightweight TinyUNet3D student with about 0.96M-parameters, approaching $\sim$145$\times$ compression with only above one point drop in performance. 
We further investigate the utility of knowledge distillation in a range of data-labeling scenarios, 
% this line can be brough back in if you have space
%experimenting with the BraTS2023 dataset  and varying labeled training data from 5\% to 100\% under identical experimental settings, 
and show that its advantage is most prominent when labeled data is scarce.
%Finally, we evaluate the effect  performance on various labeled data sizes, . 
To demonstrate generalizability of our approach, experiments are conducted on three datasets: BraTS~2023, MSD Spleen and MSD Hippocampus. The code for TinyUNet3D is publicly available at: \url{https://github.com/2802990u/tinyunet3d-dist-kd}.

\end{abstract}
\begin{keywords}
3D Segmentation, Knowledge Distillation, scarce labels, low labels domain, Deep Learning.
\end{keywords}
\section{Introduction}
\label{sec:intro}
%3D medical image segmentation labels anatomical structures, abnormalities, or lesions voxel by voxel \cite{P1,P2}, enabling accurate and timely diagnosis \cite{P3}. Traditional CNN-based models \cite{P6,cicek20163dunet} are limited by their local receptive fields, which restrict their ability to capture global context in 3D volumes. Transformers, originally developed for natural language processing \cite{P12} and general computer vision \cite{P11}, were adapted to volumetric segmentation to overcome this limitation \cite{P13}. However, their self-attention scales quadratically with input size, which makes them computationally expensive \cite{P11}. 

3D medical image segmentation can be achieved via traditional CNN-based models, but they are limited by their local receptive fields \cite{P6}. Transformer-based models can overcome this, but their self-attention scales quadratically \cite{P12}.
To capture long-range dependencies with linear-time complexity, Mamba \cite{P14}, which originates from classic state-space models \cite{P15}, was introduced to reduce training and inference overhead via a selective scanning mechanism. Although recent studies have adapted Mamba-based architectures for 2D and 3D medical image segmentation \cite{P16,P17,P18}, their full potential remains to be explored. SegMamba \cite{xing2024segmamba} and its extension SegMamba-V2 \cite{xing2025segmambav2}, key members of this growing family of State Space Model (SSM)-based models, effectively capture long-range 3D volumetric context with linear complexity, but at the cost of requiring $\sim$100M parameters and much higher FLOPs compared to other SOTA models.. Such heavy footprints hinder real-world clinical deployment, where compute and memory resources are constrained and labeled 3D training volumes remain scarce. 

Knowledge distillation (KD) \cite{hinton2015distilling} addresses these issues by transferring knowledge from a large and heavy teacher model to a compact student network. KD has shown great potential in medical image segmentation, evolving from early multimodal-to-monomodal distillation \cite{hu2020knowledge} to recent missing-modality techniques for brain tumor segmentation \cite{zhu2025bridging}. It is particularly effective when labels are scarce: soft targets from a teacher carry richer information than hard labels. In a speech-recognition experiment, a model trained on only 3\% of the data reached 57.0\% test accuracy with soft targets, against 44.5\% with hard targets~\cite{hinton2015distilling}.
A limitation of recent work in this area is that most studies evaluate KD as a teacher–student comparison conducted on one dataset with a fixed amount of training data. In addition, while Mamba-to-CNN distillation for 2D panoramic segmentation \cite{xu2026mamba} has been explored, to our knowledge, no prior work distills a Mamba/SSM-based teacher into a compact CNN student for 3D medical image segmentation. 

Our hypothesis is straightforward and two-fold: First, Mamba's soft logits can be distilled into an ultra-lightweight CNN  model for 3D medical image segmentation with acceptable accuracy; second, when labeled data is scarce, and models struggle to learn reliable decision boundaries requiring additional sources of supervision, KD should show peak advantage, diminishing as ground-truth supervision increases. 

In this paper, we directly analyze these hypotheses. A single 139M-parameter teacher~\cite{xing2025segmambav2} is distilled via correlation-based (DIST) distillation into a TinyUNet3D, a about 0.96M-parameter student designed using depthwise separable convolution ($\sim$145$\times$ smaller). The model is evaluated across three CT/MRI benchmark datasets with single- and multi-class targets. On the largest of them, BraTS~2023, we systematically vary the available training data from 5\% to 100\% under identical experimental set-up. This ensures that changes in performance are caused strictly by the amount of labeled data, rather than differences in model size, data augmentation, or dataset choice.

\noindent\textbf{Contributions.}
\begin{enumerate}
  \item \textit{3D Model Compression.} An extreme single-teacher compression ratio ($\sim$145$\times$, 139M~$\to$~0.96M-parameters) that retains over 98.8\% of the teacher’s Dice.
  \item \textit{Distillation Gain in Label-Scarce Domains.} We perform a controlled experiment that varies the training dataset size to show exactly how the size of available labeled data impacts the benefits of Knowledge Distillation, rather than just comparing performance across different datasets.
  \item \textit{Evaluation across multiple datasets.} Evaluation on three datasets spanning CT and MRI, with the data fraction sweep on BraTS~2023.
\end{enumerate}

\section{Methodology}
\label{sec:methodology}
\begin{figure*}[h] 
    \centering 
    \includegraphics[width=\textwidth, height=6.5cm, keepaspectratio]{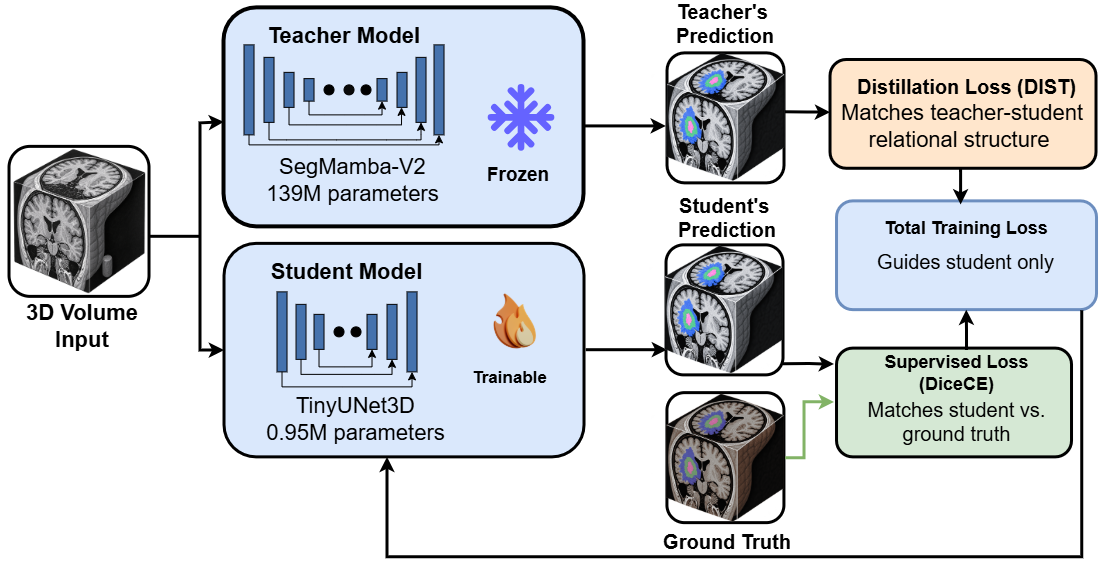} 
    \caption{Overview of the proposed framework, illustrating the distillation of knowledge from a large Mamba teacher model into an ultra-lightweight student architecture.}
    \label{fig:framework}
\end{figure*}

Our proposed pipeline leverages SegMamba v2, backbone based on Mamba state space architecture with about 139M parameters \cite{xing2025segmambav2} as the teacher and TinyUNet3D as the student. For all the datasets, the teachers are trained from scratch following an nnU-Net-style SGD setup \cite{isensee2021nnu} with a combined Dice and Cross-Entropy (DiceCE) loss. Built with depthwise separable convolutions and a maximum channel width of 64, the student model contains only about 0.96M parameters. The overall methodology is illustrated in Fig.~\ref{fig:framework}. 

During training, the student model is optimized with SGD using a DiceCE loss against the ground-truth labels plus a distillation loss computed against the frozen teacher’s predictions (Eq.~\ref{eq:total-loss}). Our distillation objective, DIST~\cite{huang2022knowledge} (Eq.~\ref{eq:dist-loss}), compares the relational structure of the teacher's and student's class-probability distributions rather than forcing the student to match the teacher's output point by point. A Pearson correlation-based distance $d(u,v) = 1 - \mathrm{corr}(u,v)$ is applied both within each voxel's class-probability vector and, per class, across all voxels (inter and intra-class relation).
\begin{equation}
  \mathcal{L} = \mathcal{L}_{\text{DiceCE}}(y_s, y_{gt}) + \lambda\,\mathcal{L}_{\text{DIST}}
  \label{eq:total-loss}
\end{equation}
\begin{equation}
  \mathcal{L}_{\text{DIST}} =
  \frac{\beta}{N}\sum_{n=1}^{N} d\big(p_s^{(n)}, p_t^{(n)}\big)
  + \frac{\gamma}{C}\sum_{c=1}^{C} d\big(p_s^{(:,c)}, p_t^{(:,c)}\big)
  \label{eq:dist-loss}
\end{equation}
where $p_t=\sigma(z_t)$ and $p_s=\sigma(z_s)$ are the teacher's and student's softmax outputs over $C$ classes, $p^{(n)}\in\mathbb{R}^C$ is one voxel's
class-probability vector ($N$ voxels total), $p^{(:,c)}\in\mathbb{R}^N$ is
one class's probability map across all voxels, and $\beta{=}\gamma{=}1.0$,
$\lambda{=}1.0$ weight the two relation terms and the overall distillation
loss.

We evaluate our model on three publicly available datasets, namely MSD Spleen, MSD Hippocampus, and BraTS2023. Among these datasets, BraTS is a large-scale brain tumor benchmark that we also use for our data fraction sweep. Specifically, we retrain the student model on 5\%, 10\%, 25\%, 50\%, and 100\% of the labeled data to see how increasing dataset size affects distillation gains. All models are implemented in PyTorch and trained on a single NVIDIA L40S GPUs (400 epochs for teachers, 250 for students) using identical pipeline for data loading, checkpointing, and evaluation. Performance is measured on the independent test sets using the mean Dice similarity coefficient, reported as the average over three random seeds for all student models (teachers are single runs).

\section{Results and Discussion}
\label{sec:results}
First, we compare our model with SOTA models and other lightweight models across all three datasets, as shown in Table~\ref{tab:results}. With about 0.96M-parameters, approximately 145$\times$ fewer than our 139M-parameter teacher, the student achieves Dice scores of 96.10, 87.59, and 88.96 on Spleen, Hippocampus, and BraTS~2023, respectively. This is within 2.4 Dice points of the best published method on each dataset, and ahead of nnU-Net and CoTr on Spleen and of SegResNet on BraTS~2023 despite their larger parameter counts. These results suggest that the lightweight design can substantially reduce model
size while maintaining competitive segmentation accuracy. 
% Requires: \usepackage{booktabs,multirow,graphicx}
% Fits ONE column (\columnwidth). Wrapper scales only if the natural width exceeds the column.
\begin{table}[t]
\centering
\caption{Comparison with SOTA methods on three datasets. Our results use a fixed held-out split, with checkpoints selected on a validation set and the test set scored once;student results are averaged over 3 seeds. Dice scores of published methods are taken from their papers (own protocols); the BraTS baselines (SegResNet, Swin UNETR-V2, SegMamba) are from Table~2 of~\cite{xing2024segmamba}. Parameters and FLOPs of the published models are computed by us (FLOPs = 2$\times$MACs, one forward pass at each model's training patch size).
}
\label{tab:results}
\savebox{\resultsbox}{%
\fontsize{8}{9}\selectfont
\setlength{\tabcolsep}{2.5pt}
\begin{tabular}{@{}l l r r r@{}}
\toprule
\textbf{Dataset} & \textbf{Method} & \textbf{Par.\,(M)} & \textbf{FLOPs\,(G)} & \textbf{Dice\,(\%)} \\
\midrule
\multirow{6}{*}{\shortstack[l]{MSD\\Spleen}}
 & UNETR~\cite{P13}        & 92.6 & 146.1 & 96.4 \\
 & nnU-Net~\cite{isensee2021nnu}      & 30 & 403.5 & 96.0 \\
 & CoTr~\cite{sota3_spleen}         & 41.9 & 561.2 & 95.4 \\
 & SegMamba-V2 (teacher) ~\cite{xing2025segmambav2}                  & 138.8 & 1566 & 96.84 \\
 & Student, no KD                   & 0.957 & 235.7 & 95.86 \\
 & \textbf{Student+DIST (ours)}     & \textbf{0.957} & \textbf{235.7} & \textbf{96.10} \\
\midrule
\multirow{6}{*}{\shortstack[l]{MSD\\Hippo.}}
 & nnU-Net~\cite{isensee2021nnu}       & 30 & 40.0 & 89.5 \\
 & ReCo-KD~\cite{sota2_hippo}       & 1.57 & 1.0 & 88.93 \\
 & Swin UNETR~\cite{hatamizadeh2022swinunetr}    & 62.2 & 636.1 & 89.19 \\
 & SegMamba-V2 (teacher) ~\cite{xing2025segmambav2}                 & 138.8 & 87.0 & 88.68 \\
 & Student, no KD                   & 0.957 & 4.9 & 87.88 \\
 & \textbf{Student+DIST (ours)}             & \textbf{0.957} & \textbf{4.9} & \textbf{87.59} \\
\midrule
%\multirow{6}{*}{AMOS-CT}
% & nnU-Net~\cite{isensee2021nnu}        & 31.4 & 404.3 & 88.87 \\
% & Swin UNETR ~\cite{hatamizadeh2022swinunetr}     & 62.2 & 637.2 & 86.37 \\
% & nnFormer~\cite{zhou2023nnformer}       & 150.5 & 358.3 & 85.63 \\
% & SegMamba-V2 (teacher) ~\cite{xing2025segmambav2}                   & 138.8 & 1567.2 & -- \\
% & Student, no KD                   & 0.957 & 300.3 & -- \\
% & \textbf{Student+DIST (ours)}     & \textbf{0.957} & \textbf{300.3} & \textbf{--} \\
%\midrule
\multirow{6}{*}{BraTS}
 & SegResNet ~\cite{P8}     & 18.8 & 1169.2 & 88.26 \\
 & Swin UNETR-V2~\cite{he2023swinunetrv2} & 72.77 & 1665.6 & 89.39 \\
 & SegMamba~\cite{xing2024segmamba}      & 67.27 & 2966.5 & 91.32 \\
 & SegMamba-V2 (teacher) ~\cite{xing2025segmambav2}            & 138.8 & 3730.7 & 88.88 \\
 & Student, no KD                   & 0.957 & 211.5 & 88.95 \\
 & \textbf{Student+DIST (ours)}     & \textbf{0.957} & \textbf{211.5} & \textbf{88.96} \\
\bottomrule
\end{tabular}}
\ifdim\wd\resultsbox>\columnwidth
  \resizebox{\columnwidth}{!}{\usebox{\resultsbox}}%
\else
  \usebox{\resultsbox}%
\fi
\end{table}

Now, we address the open question of how much accuracy gain can be attributed to
knowledge distillation across three datasets with different numbers of training cases.
Table~\ref{tab:kd_gain} answers this against the three mentioned datasets. On Spleen
dataset with 41 instances, KD gives a small gain (95.86 to 96.10, 0.24 points). On
Hippocampus dataset, which consists of 260 instances, the difference is a small drop of
0.29. On BraTS dataset with 876 training instances (out of 1,251 total, at full scale), KD
makes essentially no difference (88.95 to 88.96, 0.01 points). Hence, with the full
training labels, the KD effect stays within 0.3 points in either direction. The student
already retains 99.2\%, 98.8\% and 100.1\% of the teacher's Dice on Spleen, Hippocampus
and BraTS, respectively, with about 145$\times$ fewer parameters, so the benefit of KD
appears mainly when labels are scarce, as shown next.
\begin{table}[t]
\centering
\fontsize{9}{9}\selectfont
\caption{KD gains in Dice (\%) across three 3D medical imaging datasets, using the full training labels. Gain is Student+DIST minus No KD.}
\label{tab:kd_gain}
\setlength{\tabcolsep}{5pt}
\begin{tabular}{@{}l
  >{\centering\arraybackslash}p{1.6cm}
  >{\centering\arraybackslash}p{1.9cm}
  >{\centering\arraybackslash}p{1.3cm}@{}}
\toprule
{\bfseries Dataset} & {\bfseries No KD} & {\bfseries Student+DIST} & {\bfseries Gain} \\
\midrule
MSD Spleen
    & 95.86 & 96.10 & +0.24 \\
MSD Hippocampus
    & 87.88 & 87.59 & $-$0.29 \\
BraTS~2023 (100\%)
    & 88.95 & 88.96 & +0.01 \\
\bottomrule
\end{tabular}
\end{table}

\begin{figure*}[htbp]
    \centering
    \includegraphics[width=\textwidth, height=5.1cm]{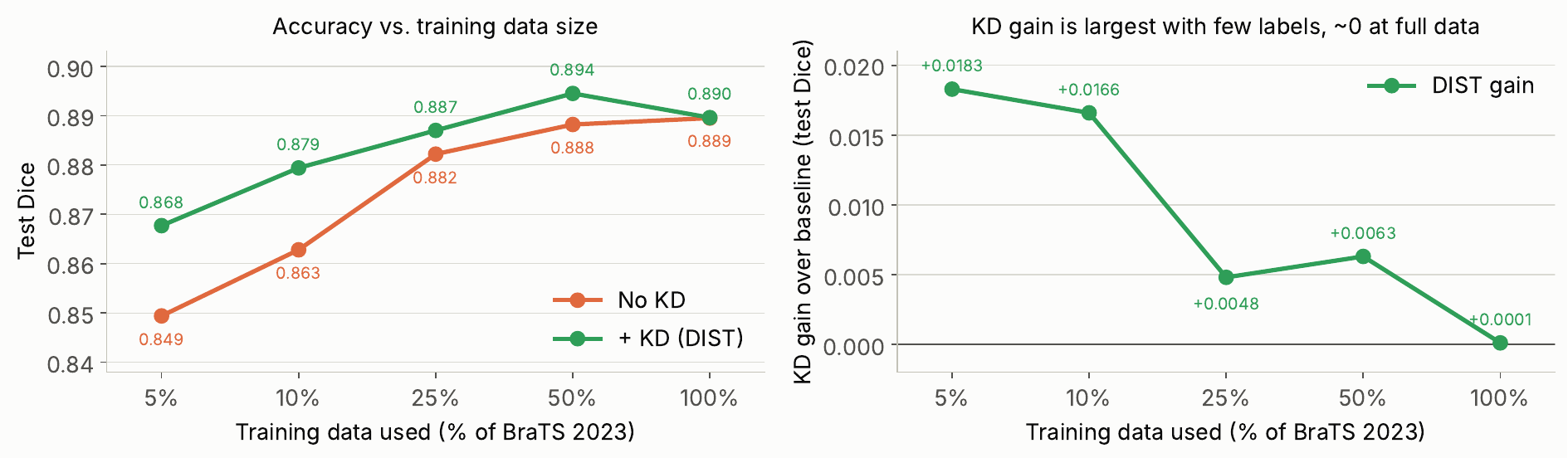}
    \caption{Effect of training data size on KD using BraTS~2023 (100\% = 876 training cases). Points are means over 3 seeds, and gains are paired by seed. Left: student test Dice (fraction) with and without DIST. Right: DIST gain over the no-KD student.
    %, which is substantially larger.
    }
    \label{fig:frac}
\end{figure*}

To further verify the effect of KD, we study its performance across different fractions of the BraTS training data (Figure~\ref{fig:frac}). With 5\% of the training data, DIST-based KD improves the student's mean Dice over the no-KD baseline by +0.018 (1.8 points), and with 10\% by +0.017 (1.7 points). Overall, the gain clearly decreases as we add more labeled data, dropping to +0.005 at 25\% and approaching 0 at 100\%, though the trend does not hold at 50\%. This is consistent with our hypothesis: distillation helps most when the student has too few labels to learn effective decision boundaries on its own, and becomes redundant once enough labeled data are available.

\begin{figure}[htbp]
    \centering
    \includegraphics[width=\columnwidth, height=4cm]{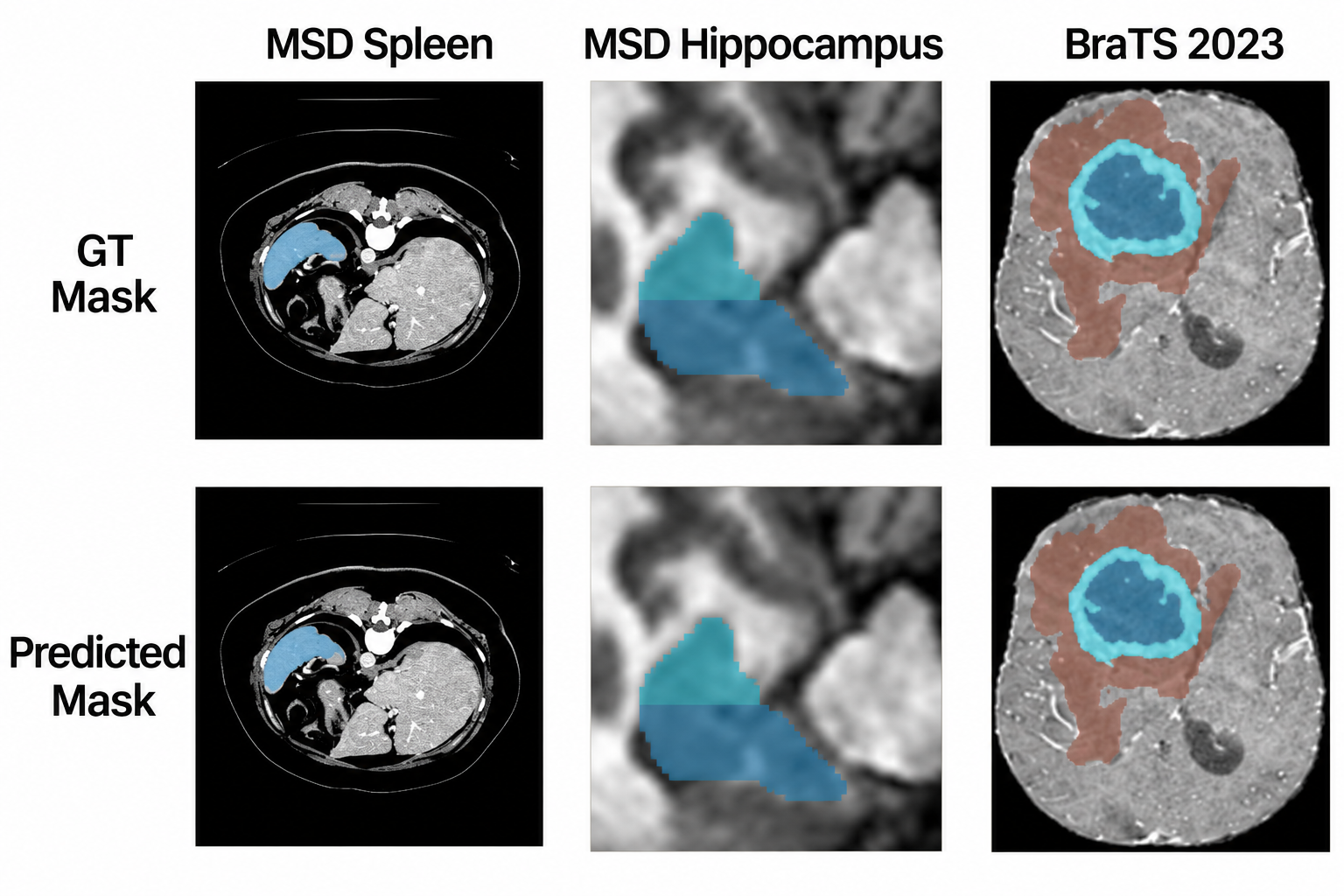}
    \caption{Qualitative segmentation results across three 3D medical imaging datasets. The first row shows the ground truth slices while the second row shows the predicted scans.}
    \label{fig:qualitative}
\end{figure}

Practically, these findings suggest using KD selectively: it should be enabled in scarce data settings where labeled instances are limited and disabled when sufficient labeled cases are available. This provides a simple adaptive strategy without additional computational overhead, as the behavior can be controlled directly through the loss weight $\lambda$. Figure~\ref{fig:qualitative} demonstrates a slice from test set for each dataset, showing the ground truth segmentation alongside the corresponding prediction of the student model.

\section{Conclusion}
\label{sec:conclusion}
In this paper, we distill a 139M-parameter SegMamba-V2 teacher into TinyUNet3D, about 0.96M-parameter CNN student that is approximately 145$\times$ smaller, for 3D medical image
segmentation. Across three CT/MRI benchmarks, the student achieved Dice scores of 96.10,
87.59, and 88.96 on Spleen, Hippocampus, and BraTS~2023, respectively. With the full
training labels, DIST-based KD changes the student's Dice by less than 0.3 points on every
dataset, so most of the student's accuracy comes from its architecture and supervised
training. The controlled data-fraction sweep on BraTS~2023 shows where KD matters: improves Dice by about 1.7 to 1.8 points with 5 to 10\% of the labels, and the gain vanishes at
100\% (0.01 points), which suggests that KD should be enabled selectively when labeled data are scarce.

In future work, we plan to extend the data fraction analysis to the other datasets and
architectures, and to evaluate the student's inference speed and memory use on
resource-constrained clinical hardware, and explore an adaptive distillation weight $\lambda$.
%We also plan to replace the fixed distillation weight $\lambda$ with one that adapts to the amount of labeled data. 

\section{Compliance with Ethical Standards}
This research study was conducted retrospectively using human subject data made available in open access: BraTS~2023 ~\cite{baid2021brats}, the MSD Hippocampus and Spleen datasets~\cite{antonelli2022msd}. Ethical approval was not required, as confirmed by the licenses attached to the open-access data.

\section{Acknowledgments}
This work has been supported by funding from Higher Education Commission (HEC), Pakistan under the ASDIS Through International Scholarships Scheme. The authors declare no conflicts of interest. 

% References should be produced using the bibtex program from suitable
% BiBTeX files (here: strings, refs, manuals). The IEEEbib.bst bibliography
% style file from IEEE produces unsorted bibliography list.
% ------------------------------------------------------------------------- 
\small
\bibliographystyle{IEEEbib}
\bibliography{refs}

\end{document}